\documentclass{article}

\usepackage{microtype}
\usepackage{graphicx}
\usepackage{subfigure}
\usepackage{booktabs} 
\graphicspath{ {figures/} }
\usepackage{amsmath}

\usepackage{hyperref}

\usepackage[accepted]{preprint_style}

\icmltitlerunning{Zero-Episode Few-Shot Contrastive Predictive Coding}

\begin{document}

\twocolumn[
\icmltitle{Zero-Episode Few-Shot Contrastive Predictive Coding:\\ Solving intelligence tests without prior training}




\begin{icmlauthorlist}
\icmlauthor{Tomer Barak}{elsc}
\icmlauthor{Yonatan Loewenstein}{elsc,others}

\end{icmlauthorlist}

\icmlaffiliation{elsc}{The Edmond and Lily Safra Center for Brain Sciences}
\icmlaffiliation{others}{Department of Cognitive Sciences, The Federmann Center for the Study of Rationality, The Alexander Silberman Institute of Life Sciences, The Hebrew University, Jerusalem}

\icmlcorrespondingauthor{Tomer Barak}{tomer.barak@mail.huji.ac.il}

\icmlkeywords{Machine Learning, meta learning, self supervised learning, intelligence tests, visual sequence prediction, contrastive predictive coding}

\vskip 0.3in
]



\printAffiliationsAndNotice{}  

\begin{abstract}
Video prediction models often combine three components: an encoder from pixel space to a small latent space, a latent space prediction model, and a generative model back to pixel space. However, the large and unpredictable pixel space makes training such models difficult, requiring many training examples. We argue that finding a predictive latent variable and using it to evaluate the consistency of a future image enables data-efficient predictions because it precludes the necessity of a generative model training. To demonstrate it, we created sequence completion intelligence tests in which the task is to identify a predictably-changing feature in a sequence of images and use this prediction to select the subsequent image. We show that a one-dimensional Markov Contrastive Predictive Coding (M-CPC$_{\text{1D}}$) model solves these tests efficiently, with only five examples. Finally, we demonstrate the usefulness of M-CPC$_{\text{1D}}$ in solving two tasks without prior training: anomaly detection and stochastic movement video prediction.
\end{abstract}

\section{Introduction}
\label{introduction}


\begin{figure}[ht!]
\begin{center}
\centerline{\includegraphics[width=\columnwidth]{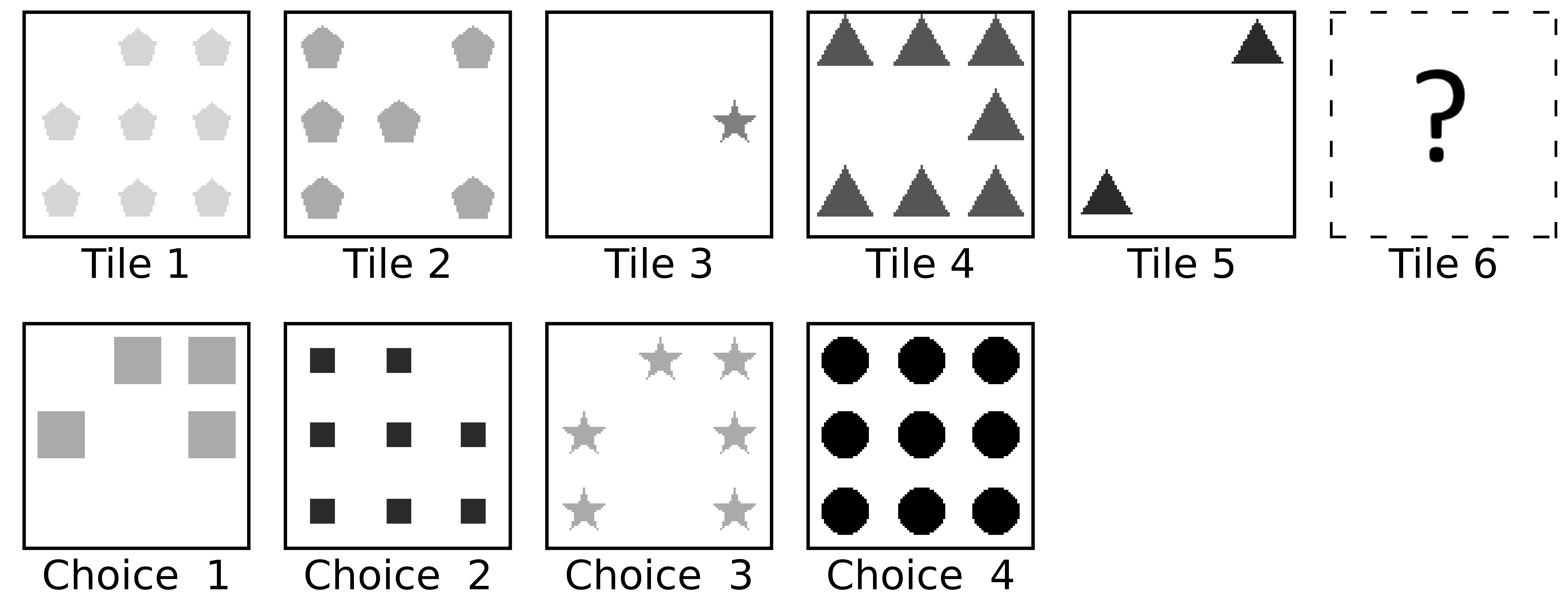}}
\caption{A sequential intelligence test. The images along the sequence become darker. The rest of their characterizing features are random. Only the 4th choice adheres to the shading rule.}
\label{fig:test_example}
\end{center}
\vskip -0.4in
\end{figure}

Changes in the world are often dominated by the dynamics of a relatively small number of latent variables. Identifying these variables is useful for making predictions. For example, in video prediction, the task is to predict an image from a sequence of its preceding images. To that goal, video prediction models often assume a small number of latent variables and learn to predict them \cite{liu_deep_2021}. However, the learning of the mapping of these latent variables to the pixel space requires the training of a generative model, which requires a large number of examples. The number of examples can be somewhat reduced if the expressivity of the latent encoder is limited \cite{kumar_videoflow_2019}, or if a simple structure is imposed on the latent variables \cite{minderer_unsupervised_2019, kim_unsupervised_2019, yang_pose_2018}.

Our focus here is on a different class of problems, in which the task is to learn a predictive latent model only and use it to evaluate the \emph{consistency} of a given image with its preceding images. Solving this problem enables the identification of incongruent images, or as we focus in this paper, the \emph{selection} of the predictable or congruent image from a set of alternative choices, rather than creating it. We argue that because no generative model training is needed, this problem can be solved using only a small number of examples. In humans, the ability to learn predictive latent variables and use them to select congruent predictions is quantified with intelligence tests, whose score is highly correlated with success in the job market and the academia \cite{sternberg_component_1977, raven_manual_1998, lohman_complex_2000, kaplan_psychological_2009, siebers_computer_2015}. Therefore, we use the framework of intelligence tests to demonstrate the data-efficiency of a latent-guided prediction by selection. 

Consider the intelligence test depicted in Fig. \ref{fig:test_example}: five ordered images are presented. The next image, the sixth, is missing. Each image is characterized by features: the number of objects, their color, shape, size, and positions. The images were constructed such that one of the features predictably changes along the sequence according to a simple deterministic rule, while the rest of the features are either constant or randomly changing. Because of the random nature of some of the features, predicting the sixth image exactly is impossible. However remarkably, humans are able to identify the image which is congruent with the sequence, out of the four alternative choices depicted in Fig. \ref{fig:test_example}, by finding the predictably changing feature without prior training.

In this paper, we use a Contrastive Predictive Coding (CPC) algorithm \cite{oord_representation_2018} which was shown useful for finding predictive latent variables \cite{anand_unsupervised_2019,henaff_data-efficient_2020,yan_learning_2020}. This self-supervised algorithm optimizes an infoNCE loss in which consecutive inputs in a sequence are regarded as positive examples and non-consecutive inputs as negative examples. Usually, to get accurate data representations, CPC models are trained over large datasets. However, we hypothesized that the selection task required for solving intelligence tests such as the one depicted in Fig. \ref{fig:test_example} are solvable by inaccurate data representations and therefore that a CPC algorithm can solve them without \emph{any} prior training. This is an extreme example of a few-shot learning in which the training is done using only the five images of the test, with zero training episodes.

\section{Intelligence Tests}
\label{sec:data}
\begin{figure}[ht]
\begin{center}
\centerline{\includegraphics[width=\columnwidth]{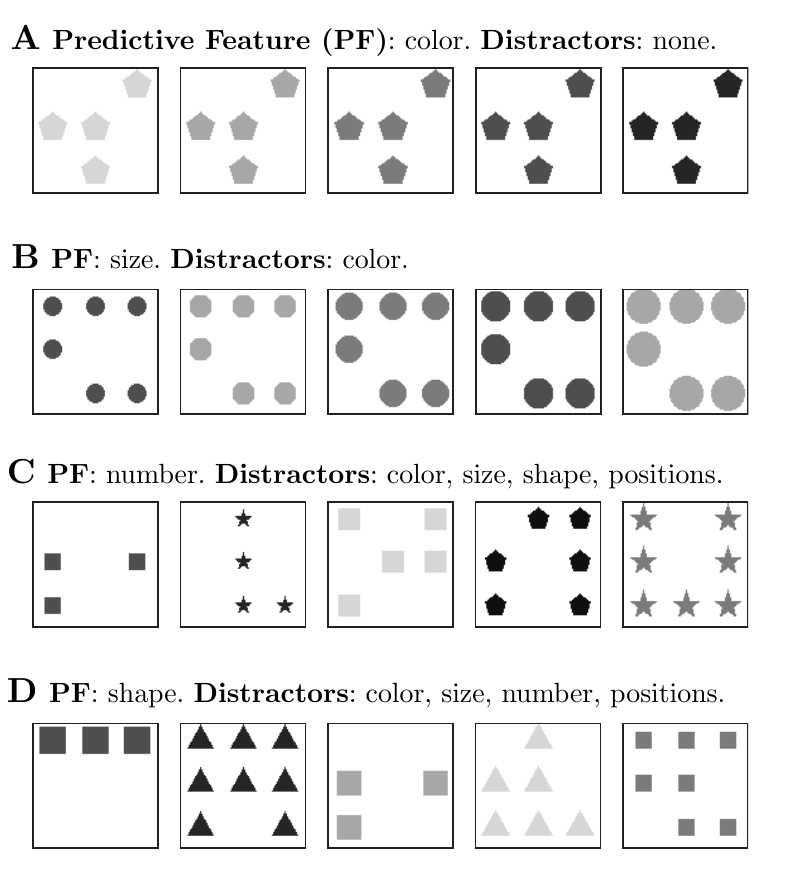}}
\vskip -0.05in
\caption{Sequences with various predictive, constant and random features. We call the random features distractors, and their number determines an intelligence test's difficulty.}
\label{fig:sequences_examples}
\end{center}
\vskip -0.2in
\end{figure}

Inspired by previous intelligence tests generating algorithms \cite{wang_automatic_2015, barrett_measuring_2018}, sequences of $K$ gray-scale images $\mathbf{x}_j$ (e.g in Fig. \ref{fig:sequences_examples}) were generated in the following way: each image included 1-9 identical objects arranged on a 3$\times$3 grid. An image was characterized by a low dimensional vector of features, $\mathbf{f}_j$ where $f^i_j$ denotes the value of feature $i$ in image $j$. We used the following five features: the number of objects in an image (possible values: 1 to 9), their shade (6 linearly distributed gray scale values), the shapes (circle, triangle, square, star, hexagon), their size (6 linearly distributed values for the shapes' enclosing circle circumference), and positions (a vector of grid positions that was used to place the shapes in order). The image $\mathbf{x}_j$ was constructed according to its characterizing features by a non-linear and complex generative function $\mathbf{x}_j=g\left(\mathbf{f}_j\right)$\footnote{Code for generating intelligence tests available on request.}.

One of the features $f^p$ predictably changed along the sequence according to a simple deterministic rule $f_{j+1}^p=u(f_j^p)$ while the other features were either constant over the images or changed randomly (values were i.i.d). We refer to the randomly-changing features as \emph{distractors} and their number is considered a measure of the difficulty of the test. After observing a sequence of $K$ images, the agent's task was to select the correct $K+1^{\text{th}}$ image from a set of $n$ optional choice images that were generated using the same generative function $g$ from the feature space. In the correct choice, $f^p$ followed the deterministic rule $f_{K+1}^p=u(f_K^p)$, whereas in the incorrect choices it did not follow that rule and was instead randomly chosen from the remaining possible values. The features that were constant or randomly changing in the sequence were also constant or changed randomly in all four choices.

\section{Markov 1D Contrastive Predictive Coding}
\label{sec:model}
Our main challenge in solving the intelligence tests is to find a latent variable that changes in a simple deterministic way along the test's five images sequence. Consider an encoder function $Z$, a predictor function $T$, two images $\mathbf{x}_a$ and $\mathbf{x}_b$ and the prediction error 
\begin{equation}
\label{eq:prediction_error}
\epsilon_{a,b}\left(Z,T\right)=\bigg(T\big(Z(\mathbf{x}_a)\big)-Z(\mathbf{x}_b)\bigg)^2    
\end{equation}
By construction, for the true encoder and predictor functions $Z^*=g^{-1}$ and $T^*=u$, $\epsilon_{a,b}\left(Z^*,T^*\right)=0$ if $a$ and $b$ are two consecutive images ($b=a+1$), and $\epsilon_{a,b}\left(Z^*,T^*\right)\neq0$ otherwise. 

The challenge is that $Z^*$ and $T^*$ are unknown. However, given a sequence of $K$ ordered images, we can approximate $Z^*$ and $T^*$ by finding $Z$ and $T$ that minimize the prediction error for consecutive images and maximize it for the non-consecutive ones. Formally, we define a contrastive infoNCE loss based on those prediction errors

\begin{equation}
\label{eq:DNNs_CPC}
    \mathcal{L}=-\frac{1}{K-1}\sum_{j=1}^{K-1}\log{
    \frac{e^{-\epsilon_{j,j+1}}}
    {\sum_{j'}{e^{-\epsilon_{j,j'}}}}}
\end{equation}
and find $Z$ and $T$ that minimize it.  

The model's encoder $Z$ and predictor $T$ are implemented by deep neural networks. Because the true predictor is one dimensional, we used a convolutional network for the encoder $Z(\mathbf{x})$ from the 100$\times$100 pixel space to a single neuron; For the predictor, $T(Z(\mathbf{x}))$, we used a residual network $T(Z(\mathbf{x}))=Z(\mathbf{x})+\Delta T(Z(\mathbf{x}))$ where $\Delta T$ is a fully-connected network. This variant of the CPC algorithm predicts a 1D latent variable based only on its most recent value. Thus, we marked it as M-CPC$_{\text{1D}}$.

\section{Results}

\subsection{Solving Tests Without Prior Training}

\begin{figure}[h]
\begin{center}
\centerline{\includegraphics[width=\columnwidth]{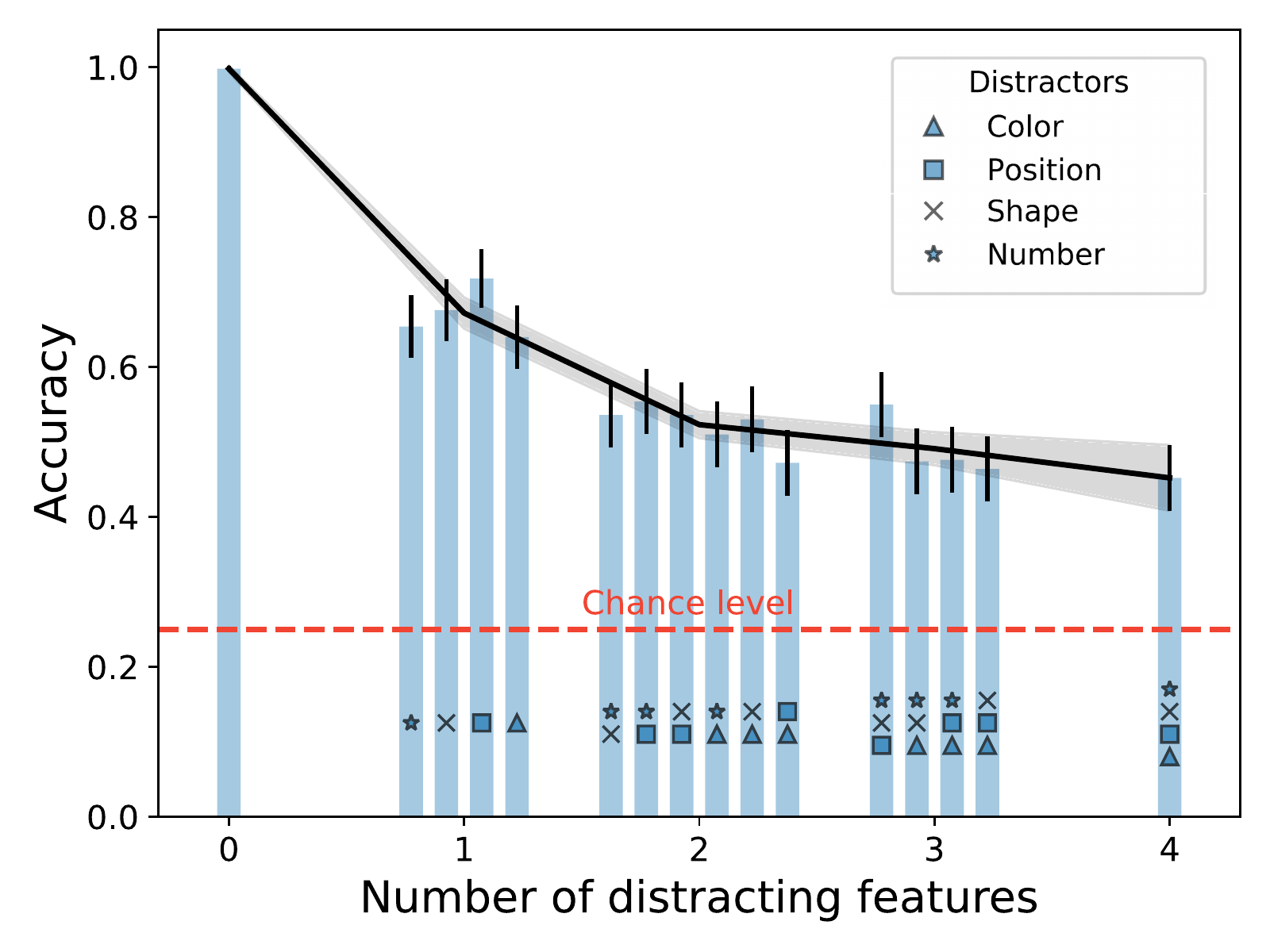}}
\vskip -0.05in
\caption{\textbf{Performance Without Prior Training.} Accuracy in 16 test conditions in which the predictive feature was the objects size that increased along the sequences. The remaining 4 features were either distractors (marked according to the legend) or constant (not marked). Performance was evaluated using $500$ randomly-generated intelligence tests (see section \ref{sec:data} for details). Error bars correspond to 95\% confidence intervals. The black line and its shade are the average accuracy per difficulty and the corresponding standard deviation.}
\label{fig:zero_episodes}
\end{center}
\vskip -0.3in
\end{figure}

To quantify the performance of M-CPC$_{\text{1D}}$ in solving intelligence tests without prior training, we applied it on a multitude of intelligence tests. For brevity, we focused in the main text on a specific predictive feature, the size of the objects, which increased monotonically throughout. Tests in which other features changed predictably are shown in the supplementary material. The 4 remaining features were either constant or randomly changing, resulting in a total of $2^4=16$ test conditions (Fig. \ref{fig:zero_episodes}). 

Each intelligence test was solved in the following way: First, we randomly initialized the networks corresponding to $Z$ and $T$. We then updated these networks' weights with a single optimization step in the direction of minimizing the loss function (Eq. \ref{eq:DNNs_CPC})\footnote{We found that a single gradient step achieved comparable results to a full minimization of the loss with more steps. We used the RMSprop optimizer with learning rate $\eta=4\cdot10^{-4}$.}. After the optimization step, we selected the choice image that had the lowest prediction error out of the four choices as the answer of this intelligence test.

Remarkably, we found that training with only the $K=5$ images that are given within the tests is sufficient for solving easy tests, as well as for achieving a substantially higher than chance performance in the more difficult tests (Fig. \ref{fig:zero_episodes}).

\subsection{Meta-Learning Sample Complexity}

To evaluate the potential benefit of prior training and obtain the sample complexity of M-CPC$_{\text{1D}}$, we tested the performance of models whose parameters are learned, rather than chosen randomly. Specifically, we performed an episode-based meta-learning, in which prior to performing a certain few-shot learning task, a model is trained on episodes (one optimization step per episode) that are random realizations of the same few-shot learning task \cite{thrun_learning_1998, sung_learning_2018}. This was done for the 16 test conditions of Fig. \ref{fig:zero_episodes}. 

\begin{figure}[ht]
\begin{center}
\centerline{\includegraphics[width=\columnwidth]{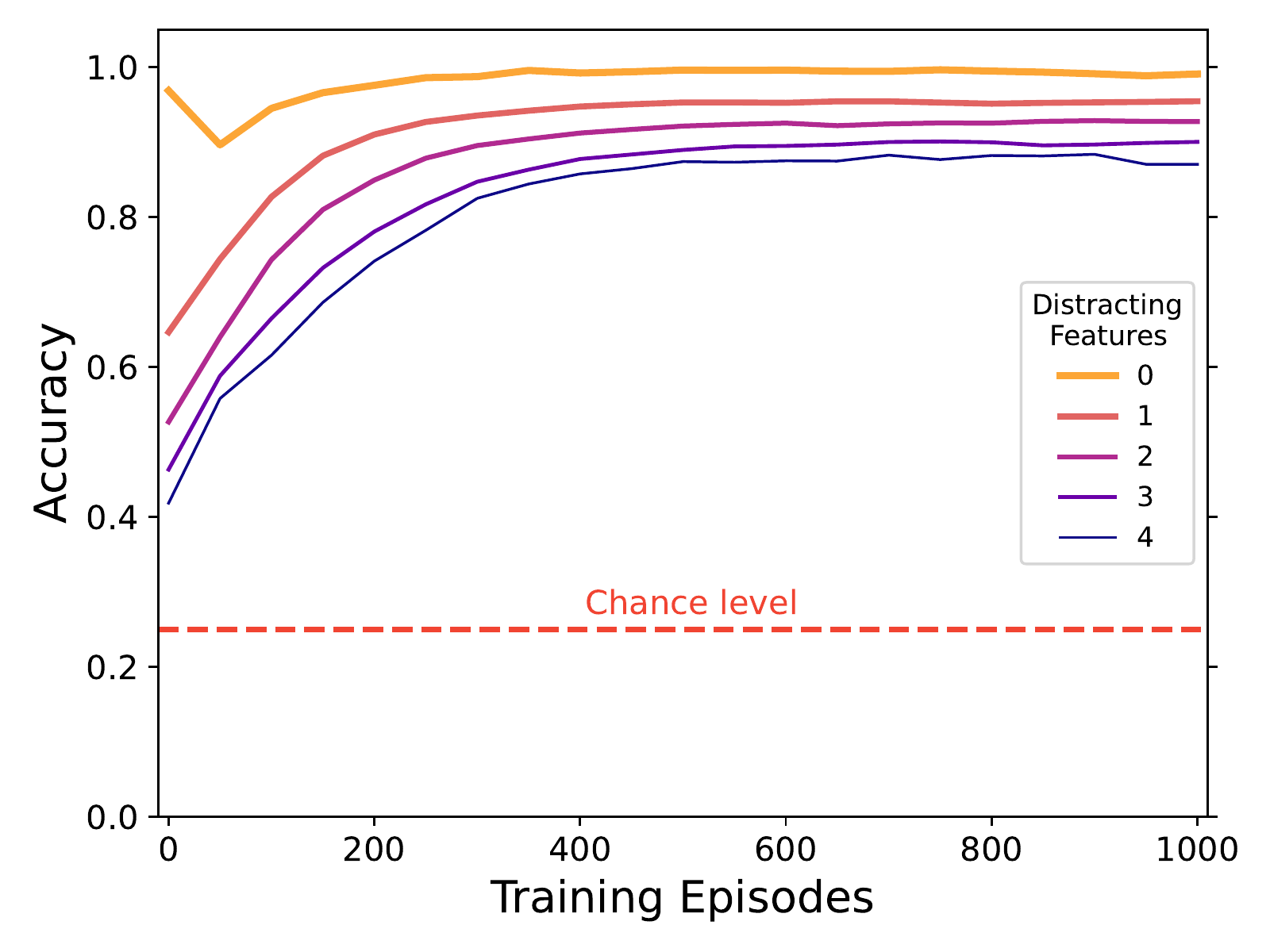}}
\vskip -0.05in
\caption{\textbf{Meta-Learning Sample Complexity.} For each of the 16 test conditions in Fig. \ref{fig:zero_episodes}, we trained the model using 1000 training episodes. Performance was evaluated every 50 training episodes. Accuracy measure corresponds to the average performance of 50 randomly-initialized models, each evaluated using 500 intelligence tests that were randomly generated from the same test condition. For clarity, performance is also averaged over test conditions that share the same number of distractors (thus, for example, 0 distractors correspond to a single test condition whereas 2 distractors depict the average over 6 test conditions).}
\label{fig:meta_learning}
\end{center}
\vskip -0.4in
\end{figure}

As depicted in Fig. \ref{fig:meta_learning}, meta-learning improved the performance of M-CPC$_{\text{1D}}$ to above 85\% accuracy in all test conditions within several hundreds of training episodes.

\subsection{Cross-Domain Generalization}

Meta-learning algorithms often overfit to the tasks they were trained on, impairing their cross-domain generalization \cite{li_learning_2017, yin_meta-learning_2020, rajendran_meta-learning_2020}. To evaluate the cross-domain generalization properties of M-CPC$_{\text{1D}}$ we extensively trained (1000 episodes) networks on intelligence tests with certain feature rules, and then tested them with intelligence tests that were characterized by other feature rules. Specifically, we trained and tested the networks using intelligence tests in which the predictive feature was either the size or the color of the objects (increasing monotonically), and the rest of the features could either be all constant (easy condition), or all distracting (hard condition). In total, we considered 4 types of intelligence tests: size-easy, size-hard, color-easy, color-hard; and we crossed between them in the training and final-evaluation stages of the model.

\begin{figure}[h]
\begin{center}
\centerline{\includegraphics[width=\columnwidth]{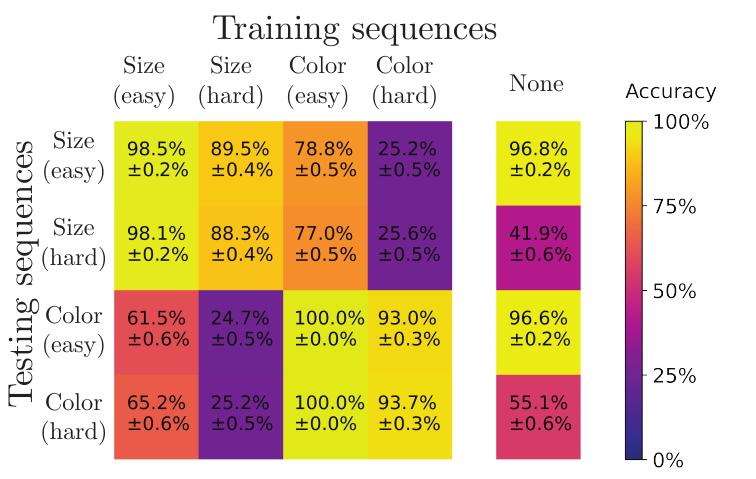}}
\vskip -0.05in
\caption{\textbf{Cross-Domain Generalization.} Randomly initiated networks $Z$ and $T$ were trained for 1000 episodes on a certain intelligence test type, and then evaluated over the 4 intelligence test types with 500 tests each. The right column presents the performance without prior training for comparison. Accuracies presented are averaged over 50 experiments and shown with 95\% confidence intervals.}
\label{fig:cross_domain}
\end{center}
\vskip -0.2in
\end{figure}

The emerging picture is interesting (Fig. \ref{fig:cross_domain}). Within the same predictive feature, we find that training using easy episodes is more effective than training with hard ones. Interestingly, after training with the easy episodes, networks' performance in the hard tests was comparable to their performance in the easy ones. These results suggest that the difference in the asymptote performances in Fig. \ref{fig:meta_learning} in the different conditions may reflects differences in the training episodes rather than differences in the difficulty of the tests. In other words, after training, the networks can solve even the most difficult tests if these tests are preceded by training using easy episodes. 

When considering training and testing on different predictive features, the picture is more complex. Training with easy episodes of one feature rule \textit{improved} performance in hard tests of the other feature rule, but was \textit{detrimental} to performance when the tests of the other rule were easy. Training with hard episodes, on the other hand, was catastrophic, bringing performance in both easy and hard tests of the other predictive feature to chance levels.

\subsection{Conclusion}

Our main result is that M-CPC$_{\text{1D}}$ can successfully solve intelligence tests without any prior training. Moreover, training the model by meta-learning can either improve or impair the performance of the model, depending on the feature alignment between the training and testing domains. These results indicate that zero-episode few-shot training can outperform trained models in environments in which domain shifts are expected, demonstrating the potential benefit of few-shot learning without any prior training.

\section{Applications}
\label{sec:applications}

\subsection{Stochastic Movement Prediction} 

To demonstrate the usefulness of M-CPC$_{\text{1D}}$, we applied it to a video prediction task. The video we chose to predict is similar to those of the Stochastic Movement (SM) dataset \cite{babaeizadeh_stochastic_2017}, in which a random shape moves in a random direction from the middle of the frame to one of its sides. It is difficult to train effective deterministic video prediction models in stochastic environments such as the SM. Therefore, stochastic video prediction models have been used to make predictions in such settings \cite{babaeizadeh_stochastic_2017, kumar_videoflow_2019}. Our approach, by contrast, was to use the data-efficiency of M-CPC$_{\text{1D}}$ to predict a video with only the first five frames of that video, without training on other stochastic videos.

\begin{figure}[h]
\begin{center}
\centerline{\includegraphics[width=\columnwidth]{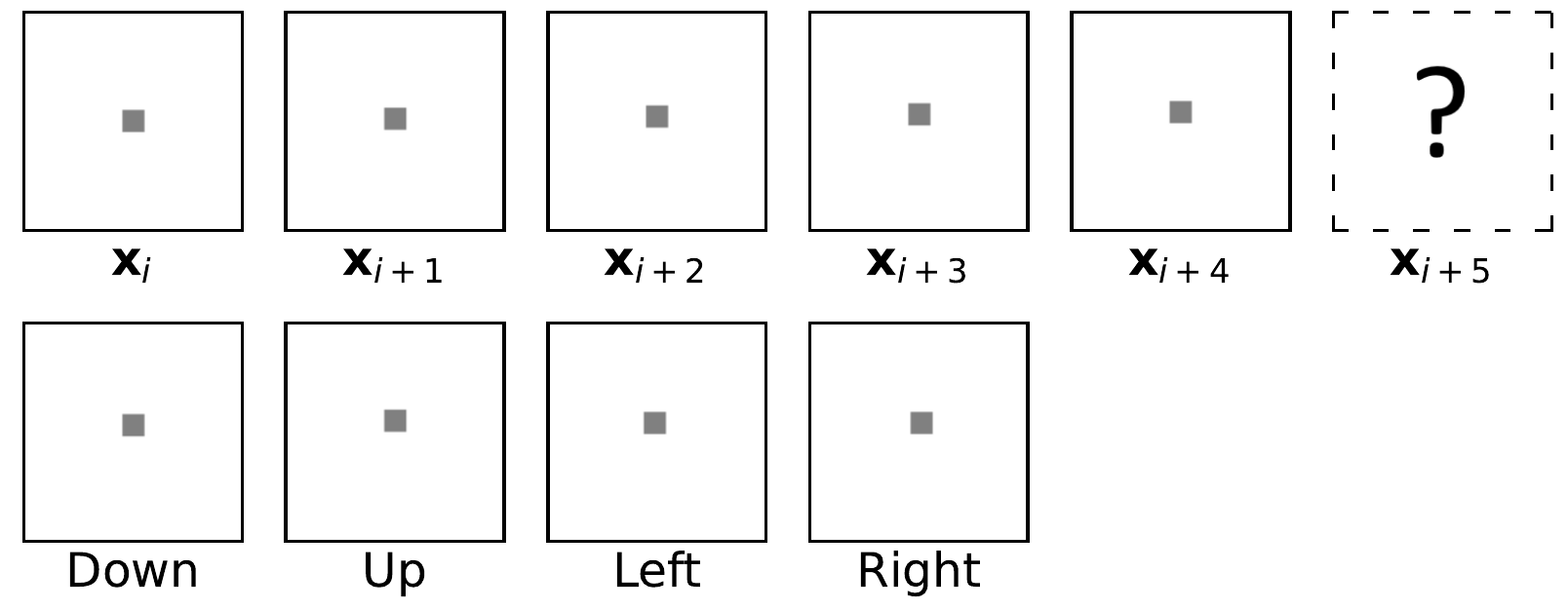}}
\vskip -0.05in
\caption{\textbf{Stochastic Movement as Intelligence Tests.} A video of a 10x10 pixel square moving upwards is predicted by treating it as an ``intelligence test''. Five consecutive frames of the video are used to predict the sixth frame, whose identity is chosen out of 4 images that are shifted by one pixel in one of the four directions relative to the last frame.}
\label{fig:Stochastic_movement1}
\end{center}
\vskip -0.1in
\end{figure}

Without loss of generality, we evaluated the video prediction ability of M-CPC$_{\text{1D}}$ with one video in which a square moved upwards (Fig. \ref{fig:Stochastic_movement1}). We optimized M-CPC$_{\text{1D}}$ on the first $K=5$ frames $\{\mathbf{x}_i\}_{i=1}^{K}$ of the video and used the latent variable to select the sixth frame $\mathbf{\tilde{x}}_{K+1}$ out of four possible motion directions, similar to the way we solved the intelligence tests. We found that based on five frames, the model can correctly predict the $K+1$ frame with a probability of $97\%\pm1\%$.

\begin{figure}[h]
\begin{center}
\centerline{\includegraphics[width=0.9\columnwidth]{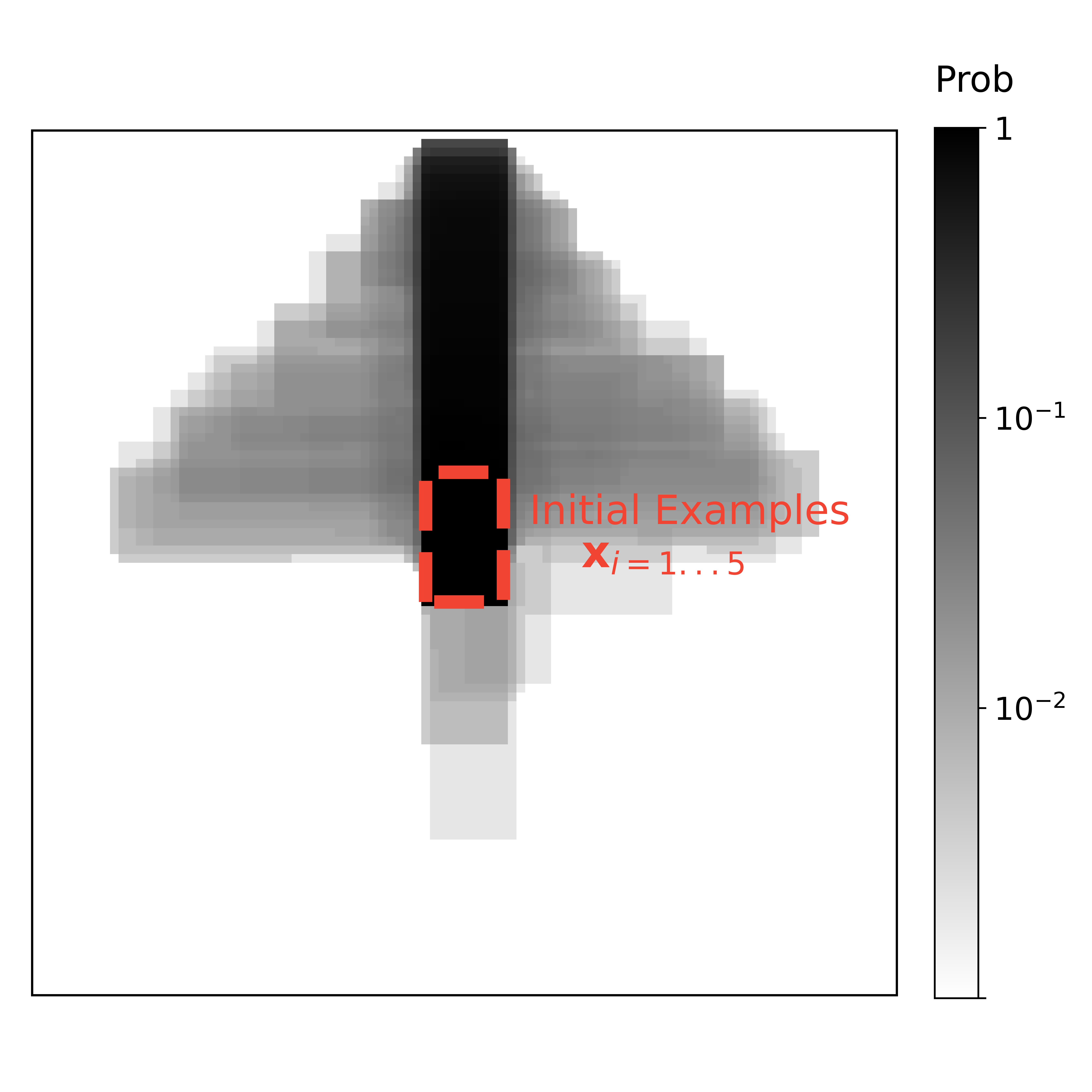}}
\vskip -0.05in
\caption{\textbf{Predicted Videos.} 500 videos obtained by video prediction with M-CPC$_{\text{1D}}$, conditioned on the $5$ initial example frames. A pixel's shade corresponds to the ratio of predicted videos, in logarithmic scale, that visited that pixel out of the 500 videos.}
\label{fig:Stochastic_movement2}
\end{center}
\vskip -0.2in
\end{figure}

To predict the next frame ($K+2$), we used the last $K-1$ frames of the original video and the predicted frame $\mathbf{\tilde{x}}_{K+1}$ to create a new sequence of length $K$ ($\{\mathbf{x}_i\}_{i=2}^{K}\cup\{\mathbf{\tilde{x}}_{K+1}\}$). We then trained new, randomly-initialized networks $Z$ and $T$, on the new sequence and used it to predict frame $K+2$, and so on. We iterated this process for 45 frames and the results are depicted in Fig. \ref{fig:Stochastic_movement2}.

As demonstrated in Fig. \ref{fig:Stochastic_movement2}, the first $K=5$ frames and the iterative process are enough for M-CPC$_{\text{1D}}$ to predict the video correctly with a high probability.

\subsection{Anomaly Detection}

To solve intelligence tests, we used a predictive latent variable to \emph{select} a congruent image from a set of alternative choice images. Predictive latent variables can also be used for anomaly detection in tasks that do not entail a selection between alternatives. Specifically, given a sequence of images, the task is to determine whether the last image is congruent or incongruent with its preceding images. 

\begin{figure}[h]
\begin{center}
\centerline{\includegraphics[width=\columnwidth]{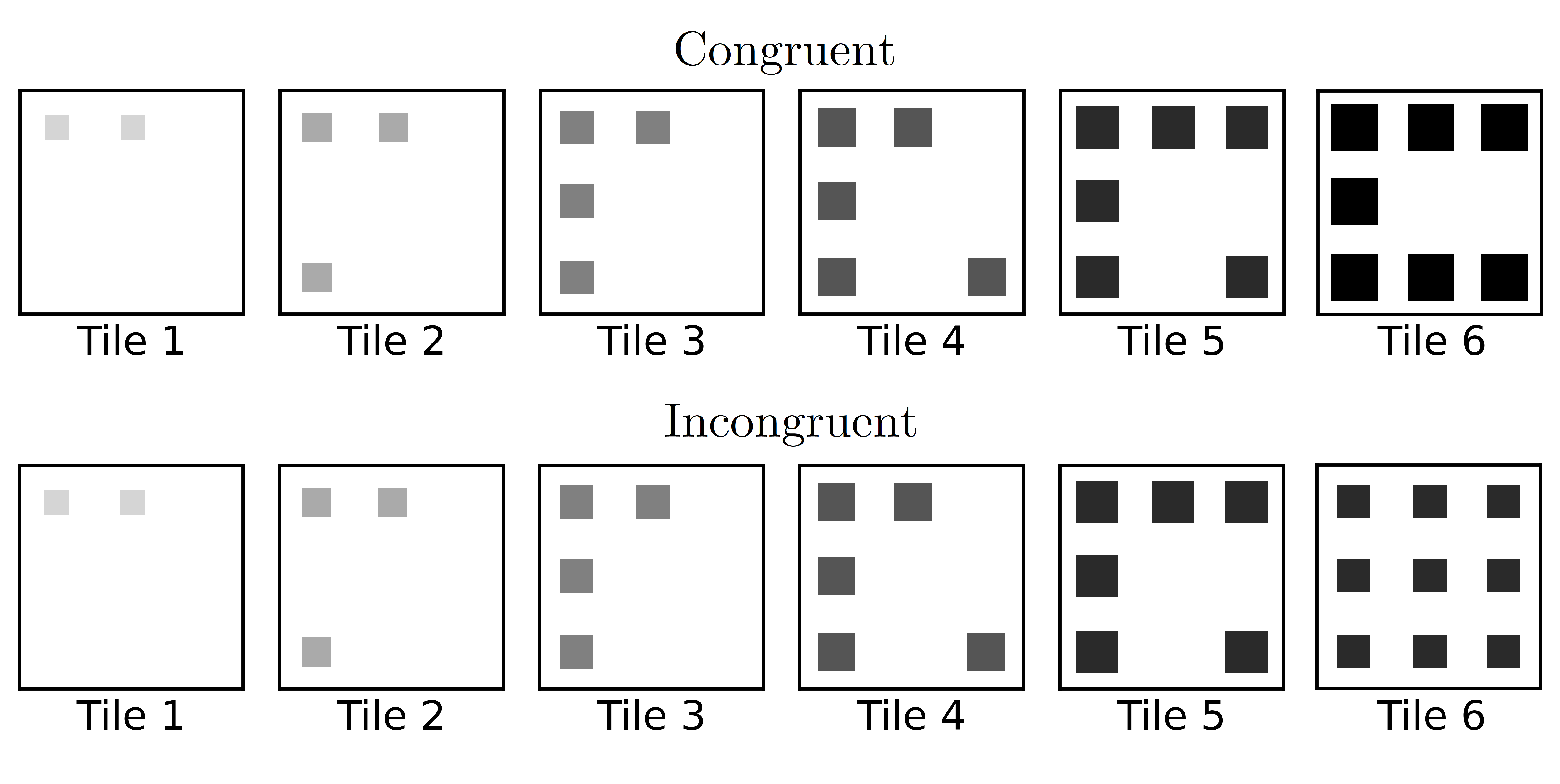}}
\vskip -0.05in
\caption{\textbf{Anomaly Detection Tests.} We provide our model a sequences of five images. Additionally, we generated two candidate images: one congruent and one incongruent with the sequence. The task of the model is to classify whether a candidate image is congruent or not, based only on the sequence, without comparing the two candidates.}
\label{fig:anomaly_detection}
\end{center}
\vskip -0.2in
\end{figure}

Consider the two sequences depicted in Fig. \ref{fig:anomaly_detection}. The task is to determine, without prior training, that there is no anomaly in the top sequence of images (tile 6 is congruent with its preceding tiles) while there is anomaly in the bottom sequence (tile 6 is incongruent with its preceding tiles). To classify the congruency of a candidate image $\mathbf{x}_c$ with a given sequence, we performed a single optimization step on the five sequence images with our loss function (Eq. \ref{eq:DNNs_CPC}). We then compared the prediction error of the candidate image $\epsilon_{K,c}$ to the average of the prediction errors of the sequence's consecutive images:
\begin{equation}
    \epsilon_{th}=\frac{1}{K-1}\sum_{j=1}^{K-1}\epsilon_{j,j+1}
\end{equation}
We use a threshold parameter $\alpha$ to classify candidate images: when $\epsilon_{K,c}>\alpha\cdot\epsilon_{th}$ we classify the candidate image as anomalous; when $\epsilon_{K,c}<\alpha\cdot\epsilon_{th}$ we classify it as congruent.

Image sequences of length $K=5$ were created such that the size, number and shade of the objects in the images increased monotonically along the sequences, while the objects' shape and the order of grid placement were constants. Two images were candidates for anomaly in each sequence: a congruent image with features that changed according to the sequence rules, and an incongruent image in which the size, number and color of the shapes did not follow the sequence rules. The shape and order of grid placement was also constant in the incongruent image. To evaluate the anomaly detection performance, we generated 500 such tests. With this classification criterion, we achieved a success rate of $85\%\pm3\%$ (taking into account false positive and misses results; chance accuracy is $50\%$). Remarkably, this result is achieved without any prior training, using only the five images of each sequence, and without relying on selection from alternatives.

\section{Conclusion}

We showed that an easy-to-train latent prediction model M-CPC$_\text{1D}$ can successfully solve prediction tasks. Specifically, we used the predictive latent variable to evaluate the consistency of an image with a sequence of preceding images by training on those preceding sequence images alone. This consistency evaluation allowed us to solve three tasks without prior training: 1) Intelligence tests, in which the task is to select the image that is most consistent with its preceding images. 2) Video prediction in which we showed that a small number of video frames, together with an iterative process, are sufficient to make some predictions in a simple video. 3) And an anomaly detection task. Common to all these examples is that reasonable performance is achieved with an extremely small number of examples. This highlights the data efficiency of latent predictions.  

It is generally argued that one main difference between human and machine learning is the amount of data required for the learning. For example, GPT-3 has been trained on hundreds of billions of words, more than three orders of magnitude more words than humans use when learning to speak \cite{hart_early_1995}. Our results demonstrate that also in machine learning, data efficiency can go a long way.    


\section{Relation to Previous Works}

\textbf{Intelligence tests solvers} - Human intelligence is often measured using intelligence tests that are similar to ours. Solving intelligence tests relies on finding relevant features in the provided examples and the rules that govern these features \cite{blum_toward_1975, siebers_computer_2015}. Traditional computational models that solved intelligence tests utilized either prior knowledge of the relevant features \cite{rasmussen_neural_2011}, knowledge about the rules that govern these features \cite{sun_deep_2018}, or both \cite{carpenter_what_1990}. Today, machine learning models are able to learn the relevant features and rules using deep artificial neural networks \cite{barrett_measuring_2018, hill_learning_2019}. However, they relied on both supervised learning and large datasets, unlike humans that seem to be able to solve such tests without prior training. Other models solved intelligence tests with unsupervised learning \cite{zhuo_solving_2020} and by meta-learning \cite{santoro_simple_2017, kim_few-shot_2020}, but they also relied on extensive prior training before solving the tests. Our work presents an ability to solve intelligence tests without any prior training.

\textbf{Contrastive learning} - Contrastive loss function are widely used for self-supervised learning \cite{,chen_simple_2020, le-khac_contrastive_2020}. One contrastive algorithm, the CPC algorithm, \cite{oord_representation_2018} is used for finding predictive latent representations, which is useful for data-efficient image recognition \cite{henaff_data-efficient_2020} and learning world-models that supports robotic object manipulation \cite{yan_learning_2020} and playing Atari games \cite{anand_unsupervised_2019}. Compared with these works, which relied on extensive training, we trained M-CPC$_{\text{1D}}$ with only five images. This was possible because we used a 1D latent variable feed-forward encoder rather than a higher-dimensional recurrent network encoder. 

\textbf{Video prediction models} - The video prediction task is useful for various down-stream applications such as representational learning and model-based reinforcement learning. Therefore, many deep learning models were developed to solve this task \cite{oprea_review_2020}. State-of-the-art models utilize a latent prediction models to solved the task \cite{minderer_unsupervised_2019, kim_unsupervised_2019, yang_pose_2018, lee_revisiting_2021}. These models also required generative models, which needed training as well. A stochastic model for the latent variable has been shown useful for video generation \cite{kumar_videoflow_2019, franceschi_stochastic_2020, babaeizadeh_stochastic_2017}. However, as we show here, a deterministic latent model is sufficient for the selection task even in stochastic environments, which allows for better data efficiency. 

\textbf{Relation networks and meta-learning} - M-CPC$_{\text{1D}}$ is similar to Relation Network (RN) \cite{sung_learning_2018}, a model which utilized episode based meta-learning for classifying examples by the abstract relation between them. While RN can learn general abstract relations between inputs, our model is inductively biased for learning a future-directed residual relation between consecutive latent representations. As a result of this and the fact that the task is a selection task, RN models require a large number of episodes for training while the intelligence tests can be solved without any prior training.

\bibliography{ICML2022}
\bibliographystyle{icml2021}

\renewcommand{\thesection}{S\arabic{section}}
\renewcommand{\thefigure}{S\arabic{figure}}

\newsavebox{\imagebox}

\title{Supplementary Materials}
\date{}

\maketitle

\section{Results of other predictive features}

\begin{figure}[ht!]
\begin{center}
\centerline{\includegraphics[width=\columnwidth]{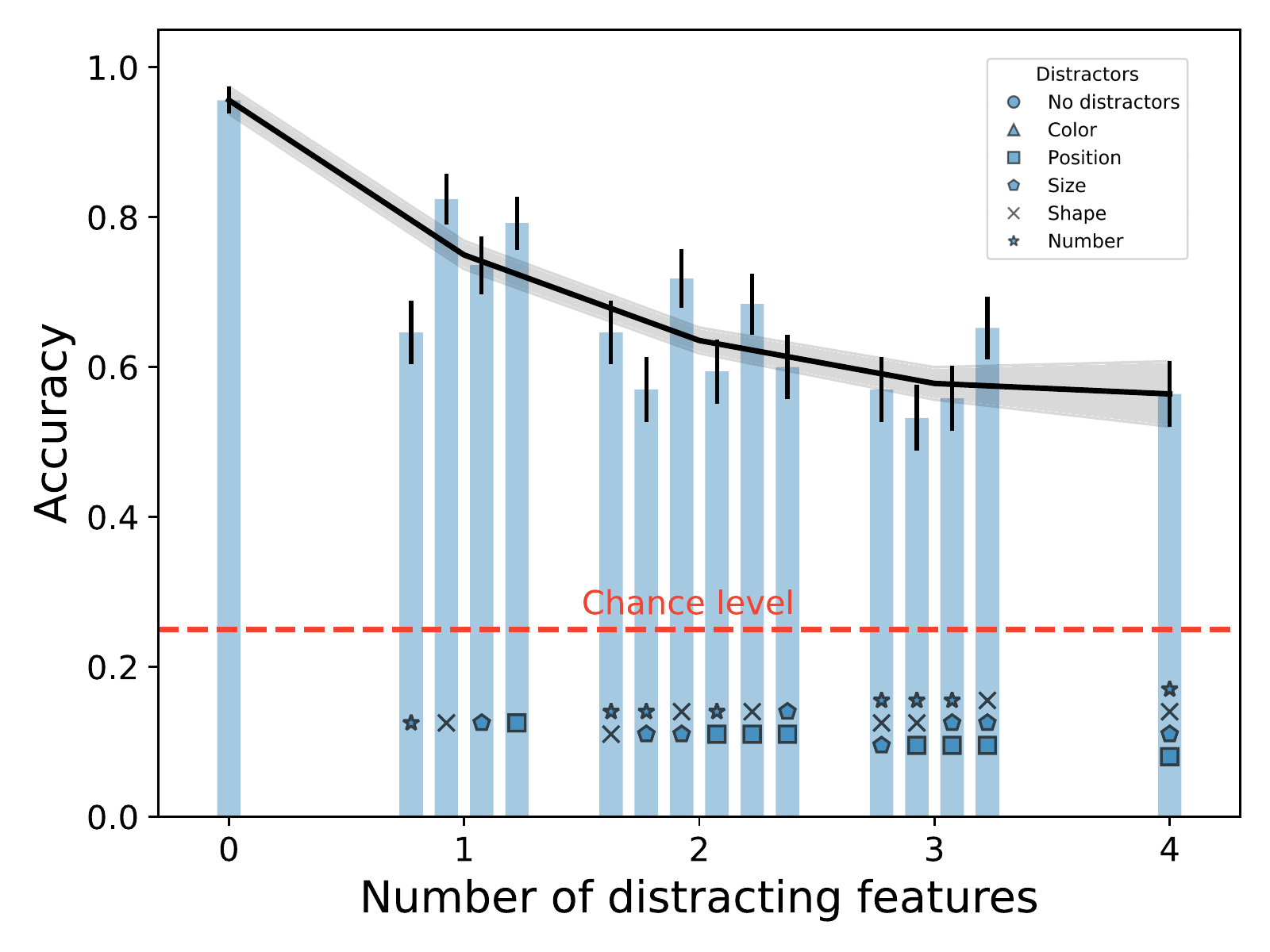}}
\caption{\textbf{Predictive Feature: color.} Zero-episodes performance when the color of the shapes became darker along the sequences.}
\label{fig:color}
\end{center}
\end{figure}

\begin{figure}[ht!]
\begin{center}
\centerline{\includegraphics[width=\columnwidth]{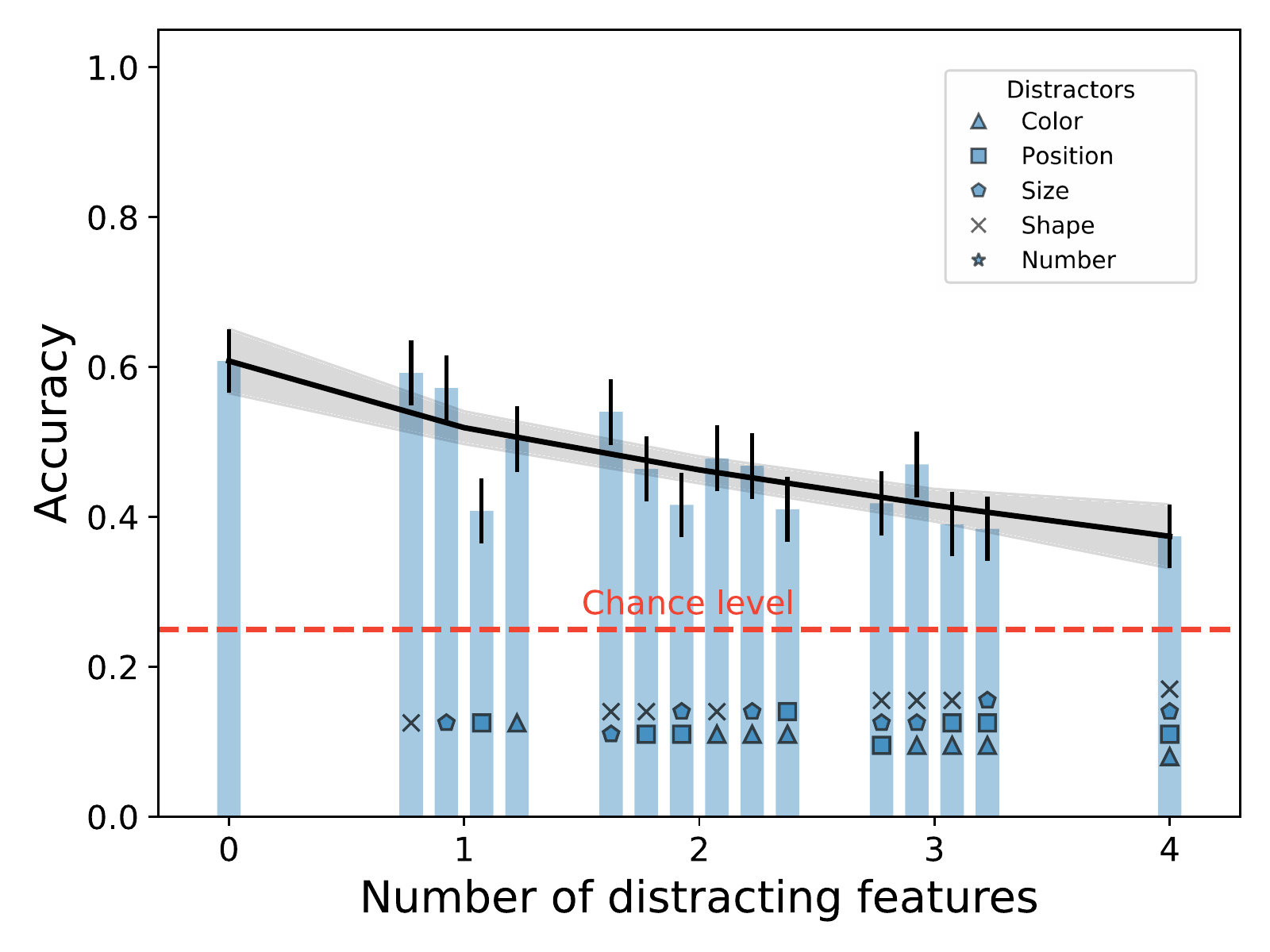}}
\caption{\textbf{Predictive Feature: number.} Zero-episodes performance when the number of shapes increased monotonically along the sequences.}
\label{fig:number}
\end{center}
\end{figure}

\begin{figure}[ht!]
\begin{center}
\centerline{\includegraphics[width=\columnwidth]{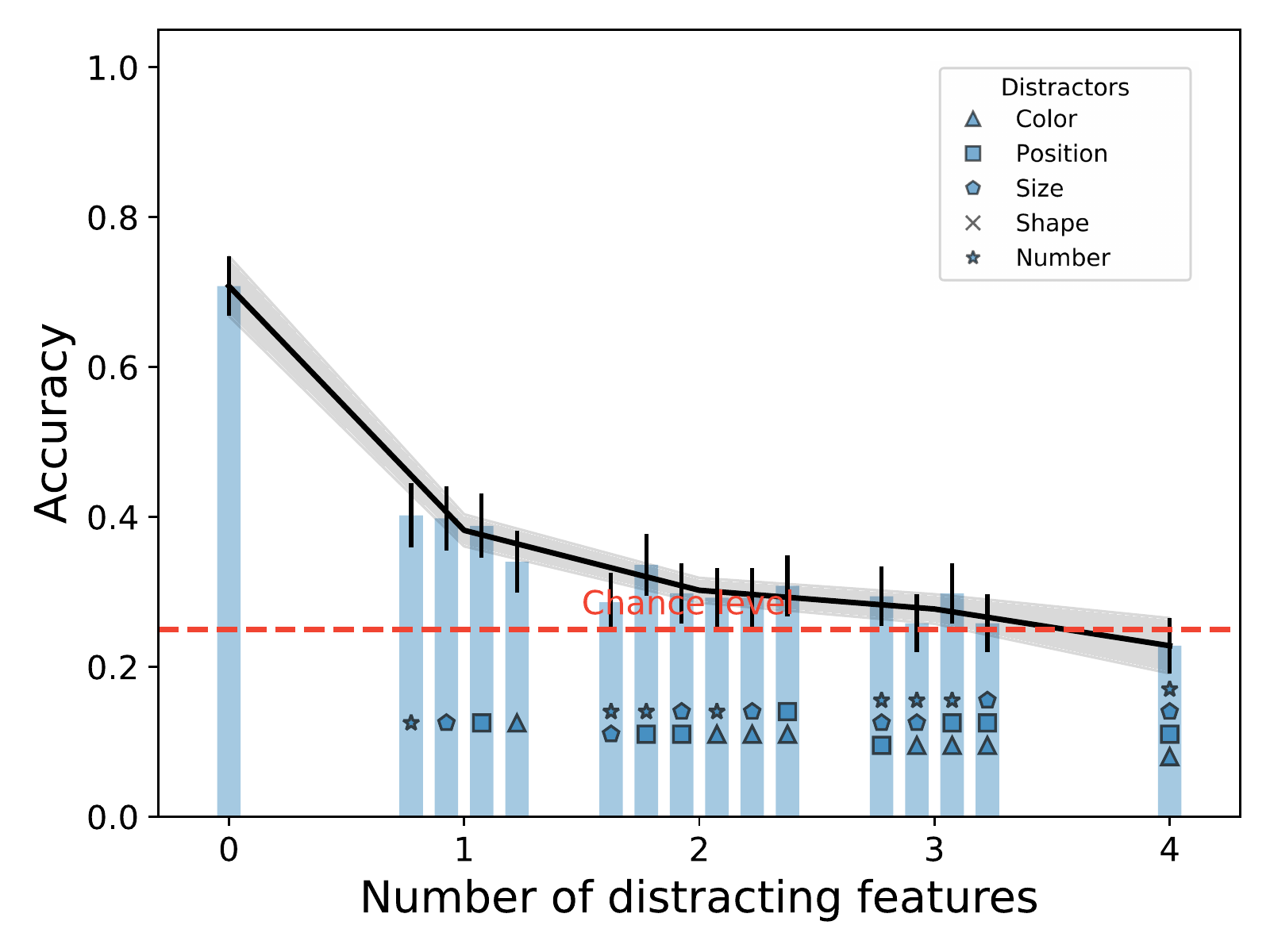}}
\caption{\textbf{Predictive Feature: shape.} Zero-episodes performance when the shapes alternated between a triangle and a square along the sequences.}
\label{fig:shape}
\end{center}
\end{figure}

\end{document}